\documentclass[10pt,twocolumn,letterpaper]{article}

\usepackage[pagenumbers]{cvpr} 

\usepackage{microtype}

\renewcommand{\paragraph}[1]{\vspace{.4em}\noindent\textbf{#1.}}
\usepackage{soul}
\setuldepth{foobar}

\usepackage{bm}
\usepackage{amsmath}
\usepackage{amssymb}
\usepackage{amsthm}

\usepackage{multirow}
\usepackage{multicol}
\usepackage{makecell}

\usepackage{adjustbox}
\usepackage{booktabs}
\usepackage{threeparttable}
\usepackage{tabularx}
\usepackage{subcaption}
\usepackage[table]{xcolor}

\usepackage{algorithm}
\usepackage{algorithmicx}
\usepackage[noend]{algpseudocode}
\usepackage{bbding}

\usepackage{pifont}
\usepackage{mathtools}
\usepackage{thmtools}
\usepackage{thm-restate}

\newcommand{\papertitle}{IC-World: In-Context Generation for Shared World Modeling}
\newcommand{\methodname}{IC-World}

\definecolor{cvprblue}{rgb}{0.21,0.49,0.74}
\usepackage[pagebackref,breaklinks,colorlinks,allcolors=cvprblue]{hyperref}

\def\paperID{5432} 
\def\confName{CVPR}
\def\confYear{2026}

\title{\papertitle}

\author{
    Fan Wu\textsuperscript{1} \quad
    Jiacheng Wei\textsuperscript{1} \quad
    Ruibo Li\textsuperscript{1} \quad
    Yi Xu\textsuperscript{2} \quad
    Junyou Li\textsuperscript{3} \quad
    Deheng Ye\textsuperscript{3} \quad
    Guosheng Lin\textsuperscript{1}\thanks{Corresponding author.} \\
    \textsuperscript{1}Nanyang Technological University \quad \textsuperscript{2}Goertek Alpha Labs \quad
    \textsuperscript{3}Tencent \\
    {\tt\small fan011@e.ntu.edu.sg \quad gslin@ntu.edu.sg} \\
    {\small \textcolor{magenta}{\texttt{https://github.com/wufan-cse/IC-World}}}
}

\begin{document}

\twocolumn[{
\begin{@twocolumnfalse}
\maketitle
\begin{center}
\setlength{\belowcaptionskip}{-1pt}
    \captionsetup{type=figure}
    \includegraphics[width=0.95\linewidth]{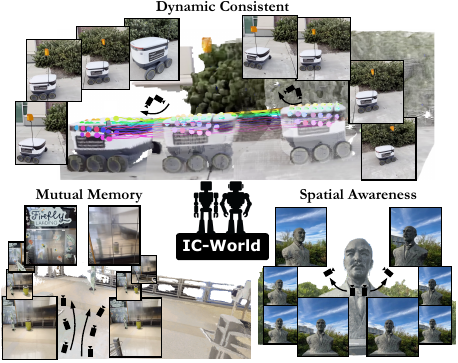}
    \captionof{figure}{\textbf{\methodname}~is a shared world modeling framework, which aims at generating $N$ shared world videos from $N$ input images, respectively, where all input images are snapshots of the same underlying world. 
    Here we present cases of two generated videos with two unconstrained camera trajectories.
    This figure illustrates three core capabilities of~\methodname: (1) \textbf{Dynamic consistency} - object motions remain coherent across different views. (2) \textbf{Mutual memory} - scenes reappear consistently, as shown by the complete words above the gate generated in the left view. (3) \textbf{Spatial awareness} - both foreground objects and background scenarios exhibit strong geometry consistency.}
    \label{fig:teaser}
\end{center}
\end{@twocolumnfalse}
}]

\begin{abstract}
Video-based world models have recently garnered increasing attention for their ability to synthesize diverse and dynamic visual environments.
In this paper, we focus on shared world modeling, where a model generates multiple videos from a set of input images, each representing the same underlying world in different camera poses.
We propose~\methodname, a novel generation framework, enabling parallel generation for all shared world input images via activating the inherent in-context generation capability of large video models.
We further finetune~\methodname~via reinforcement learning, Group Relative Policy Optimization, together with two proposed novel reward models to enforce scene-level geometry consistency and object-level motion consistency among the set of generated videos.
Extensive experiments demonstrate that~\methodname~substantially outperforms state-of-the-art methods in both geometry and motion consistency.
To the best of our knowledge, this is the first work to systematically explore the shared world modeling problem with video-based world models. 
\end{abstract}

\vspace{-16pt}
\section{Introduction} 

With the recent advances in video-based world models~\cite{hong2022cogvideo,HaCohen2024LTXVideo,wan2025,kong2024hunyuanvideo,xu2024easyanimate,chen2024videocrafter2,xing2023dynamicrafter}, shared world modeling has emerged as an important extension.
We define shared world modeling as the task of generating a set of videos that are both spatially and temporally consistent, where each video corresponds to a distinct input image of the same underlying world and unlimited camera trajectories, as shown in~\Cref{fig:teaser}.
This problem forms a foundation of numerous critical applications, including multiplayer video gaming and multi-robot coordination simulation, all of which involve multiple independent cameras that must coherently represent a single shared world.


Existing video-based world models contrast with this task, as they typically generate videos from a single viewpoint, i.e., image-to-video (I2V).
Directly extending them to this task remains non-trivial, since sequentially and independently generating videos of the shared world input images would introduce complex dependencies across generated videos to ensure spatial and temporal consistency.
Moreover, hallucination effects~\cite{sahoo2024comprehensive,chu2024sora} observed in these models further amplify inconsistencies between generated videos.
Another lines of work focus on domain-specific world modeling.
Multi-view video generation for autonomous driving~\cite{russell2025gaia,zhao2025drivedreamer,zheng2024genad} aims to generate consistent cross-camera views to simulate the views of ego-vehicles.
4D video generation~\cite{wang20254real} aims to generate consistent object-centric videos with continuous viewpoints.
However, these works either use a preset multi-camera system or focus on object-centric generation, limiting their usage in advanced world modeling applications, such as multiplayer video gaming, which requires scene-level generation with multiple free cameras.


To address shared world modeling, we introduce~\methodname, a novel generation framework that leverages the inherent in-context generation capability~\cite{fei2024video} of large video models to generate shared world videos.
As shown in~\Cref{fig:teaser}, our method effectively performs shared world modeling, exhibiting mutual memory across generated views.
To further enhance consistency, we finetune~\methodname~with reinforcement learning, Group Relative Policy Optimization (GRPO), and equip it with two novel reward models designed to ensure scene-level geometry consistency and object-level motion consistency across generated videos.
Extensive experiments demonstrate that the proposed framework significantly surpasses state-of-the-art video-based world modeling methods on our benchmark and exhibit strong visual quality as shown in~\Cref{fig:teaser}.

In summary, our contributions are as follows:
\begin{enumerate}
    \item To the best of our knowledge, this is the first work to formulate shared world modeling as a new setting for video-based world modeling and to establish a comprehensive benchmark with well-designed evaluation metrics.
    \item We proposed~\methodname, a novel instructional generation framework to activate the inherent in-context generation capability of large video models to generate multiple shared-world videos in a parallel manner.
    \item We introduce GRPO-based reinforcement finetuning with two novel reward models to further enforce scene-level geometry and object-level motion consistency.
\end{enumerate}

\section{Related Works}

\paragraph{Video-based world models}
The emergence of Sora has further highlighted the potential of large video generation models as world simulators.
Recent advancements have primarily focused on text-to-video (T2V)~\cite{chen2023videocrafter1,chen2024videocrafter2,li2024t2v}, and
image-to-video (I2V)~\cite{hong2022cogvideo,yang2024cogvideox,wan2025,HaCohen2024LTXVideo,xu2024easyanimate,agarwal2025cosmos}.
However, conventional I2V methods struggle to address the shared world modeling, as they fail to guarantee that all generated videos depict the same underlying world.
Other extension tasks of video-based world modeling, including domain-specific multi-view video generation~\cite{russell2025gaia,zhao2025drivedreamer,zheng2024genad,wu2024drivescape} and general 4D video generation~\cite{wang20254real,wu2025cat4d,xie2024sv4d}, are either defining fixed camera systems or limited to object-centric videos, which are less generalizable compared with our problem setting.


\paragraph{Instructional in-context generation}
Recent studies in image generation~\cite{huang2024context,zhang2025context,wu2025qwenimagetechnicalreport,deng2025bagel} have revealed that large-scale diffusion transformers (DiTs) inherently possess in-context generation capabilities, which can be activated through appropriately designed instructional prompts.
IC-LoRA~\cite{huang2024context} demonstrates that existing text-to-image DiTs are capable of performing in-context generation without additional fine-tuning. 
Similarly, IC-Edit~\cite{zhang2025context} introduces an instruction-based framework that exploits the latent comprehension and generation abilities of large-scale DiTs for image editing.
Beyond text-to-image, Fei et al.~\cite{fei2024video} extend this to video models, demonstrating the remarkable adaptability of the idea.
In this paper, we extend this inherent in-context capability to shared world modeling and propose a novel framework to generate consistent videos. 

\paragraph{Reinforcement finetuning for video models} 
Reinforcement learning has recently emerged as an effective paradigm for optimizing large-scale generative models~\cite{liu2025videodpo,yuan2024self,zhang2024onlinevpo,xue2025dancegrpo}, particularly for enhancing reasoning and decision-making capabilities. 
DeepSeek-R1~\cite{guo2025deepseek} introduces Group Relative Policy Optimization (GRPO), which leverages verifiable reward signals to estimate relative advantages among model responses, thereby substantially improving reasoning performance.
Following this advancement, GRPO-based fine-tuning has been successfully extended to a variety of multimodal domains, including video understanding~\cite{feng2025video,wang2025time}, multi-image grounding~\cite{bai2025univg}, and visual generation~\cite{fang2025got,xue2025dancegrpo}.
These developments collectively demonstrate the versatility of GRPO in optimizing the complex visual generation task.
Building upon this, we design two novel reward models to enforce scene-level geometry consistency and object-level motion consistency across the generated videos for shared world modeling.

\section{Methodology}

\paragraph{Problem setup} 
Shared world modeling aims to output a set of $N > 1$ videos $\mathcal{V}=\{v_i\}_{i=1}^N$ conditioned on a corresponding set of $N$ input images $\mathcal{I}$ and a shared text prompt $c$ without a constrained camera trajectory for each input image.
The input images capture the same world at the same time from different camera views; in other words, the input images are multi-view and fixed-time snapshots of an underlying world described by $c$, including its static scenes and dynamic objects.
In this study, we focus on the simplest but non-trivial case with $N = 2$ since we observe that even this setting remains challenging for existing methods demonstrated in our experiments.

\subsection{Preliminary}
DanceGRPO~\cite{xue2025dancegrpo} is a reinforcement learning framework that extends GRPO to visual generation tasks.
The central idea is to reinterpret the denoising process as a Markov Decision Process (MDP), where each timestep represents a state transition and the reward of the final output. 
Formally, for a given prompt $c$, the state at denoising timestep $t$ is defined as $s_t = (c, t, z_t)$, where $z_t$ denotes the latent at $t$. 
The policy model $\pi_\theta$, which is the video model with parameters $\theta$, will generate $M>1$ outputs $\mathcal{O}=\{o_i\}_{i=1}^M$ and governs the transition from $z_t$ toward predicting lower-noise latent $z_{t-1}$ with the probability policy function:
\begin{equation}
\pi_\theta(z_{t-1,i} \mid s_{t,i}) \coloneqq  p(z_{t-1,i} \mid z_{t,i}, c), i\in[1,M],
\end{equation}
where each output $o_i$ contains $\{v_i\}_{i=1}^N$ videos of $N$ input images.
Learned or rule-based reward model will then compute reward $\{r_i\}_{i=1}^M$ for each output.
Then the training of GRPO maximizes the following objective $L(\theta)$:
\begin{equation}
\label{eq:grpo_objective}
\begin{split}
L(\theta) &= \mathbb{E} \left[ \min\!\left(\rho_{t,i} A_i,\,\text{clip}(\rho_{t,i}, 1-\epsilon, 1+\epsilon) A_i \right)\right] ,
\end{split}
\end{equation}
where 
$\rho_{t,i} = 
\frac{
\pi_{\theta}(z_{t-1,i}\mid s_{t,i})
}{
\pi_{\theta_{\text{prev}}}(z_{t-1,i}\mid s_{t,i})
}$ is the policy ratio between the previous sample and the current sample, and the group-normalized advantage is computed as:
\begin{equation}
\label{eq:advantage}
A_i = \frac{r_i - \text{mean}(\{r_j\}_{j=1}^M)}{\text{std}(\{r_j\}_{j=1}^M)}.
\end{equation}
This normalization allows each sample’s contribution 
to be evaluated within its group, enhancing training stability.





\begin{figure}[t]
    \centering
    \includegraphics[width=\linewidth]{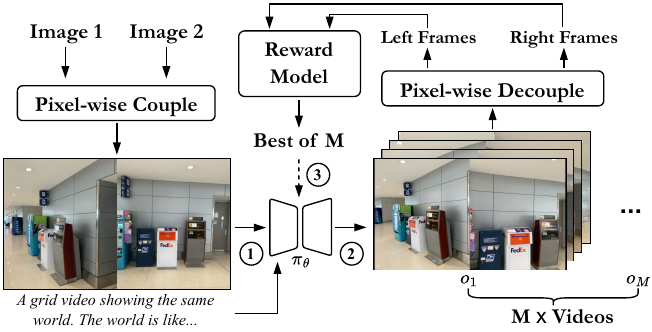}
    \caption{\textbf{Overview of~\methodname.} Here we present the case of $N=2$. (1) The input images are first concatenated pixel-wise, accompanied by an in-context prompt, to perform I2V generation. (2) The model $\pi_\theta$ then generates $M$ candidate videos conditioned on the same input. (3) Each candidate video is decoupled pixel-wise into sub-videos for reward calculation. Then we find the best candidate to calculate gradients for finetuning.}
    \label{fig:methodology}
\end{figure}

\subsection{In-Context Generation}


Shared world modeling aims to generate $N$ videos given a shared text prompt $c$ and $N$ input images.
Conventional I2V models generate each video sequentially and independently, making it difficult to ensure the dependencies required by output videos, where all videos share the same world.
This inconsistency is further amplified by the inherent hallucination tendencies of I2V models.
To address this limitation, we introduce a minor yet effective modification to conventional video diffusion transformers. 
Specifically, we design an in-context prompt template for shared world modeling as follows:
\textit{``A grid video showing the same, synchronized world captured from different camera poses. The world is like: \textless WORLD\textgreater."}
\textless WORLD\textgreater{} denotes a placeholder that is substituted with the textual description of the underlying world.
Alongside the prompt, we downscale each input image and combine them pixel-wise into a single grid image as shown in~\Cref{fig:methodology}, in a way to serve as the model input and perform parallel generation for all input images.
After inference, we decouple the generated large grid videos into $N$ small sub-videos pixel-wise.
With this generation framework, the inherent capability of large video models as the world's in-context generator can be activated to output consistent videos.



\begin{figure}[t]
    \centering
    \begin{subfigure}[t]{\linewidth}
        \centering
        \includegraphics[width=\linewidth]{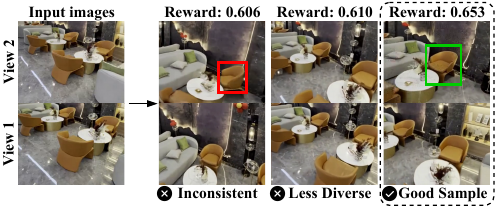}
        \caption{\textbf{Geometry consistency} under the static scene and dynamic camera.}
        \label{fig:reward_score_visualize1}
    \end{subfigure}
    
    \begin{subfigure}[t]{\linewidth}
        \centering
        \includegraphics[width=\linewidth]{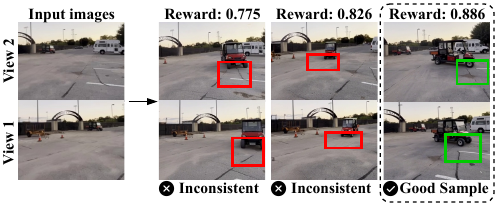}
        \caption{\textbf{Motion consistency} under the dynamic scene and static camera.}
        \label{fig:reward_score_visualize2}
    \end{subfigure}
    \caption{\textbf{Reward visualization.} Our proposed reward models are able to choose the best candidate that aligns well with human visual perception. Better view in zoom-in mode.}
    \label{fig:reward_score_visualize}
\end{figure}

\subsection{Shared World Reward Models}
To further enhance consistency among generated sub-videos, we proposed two reward models for consistency on scene-level geometry and object-level motion.
For clarity, we illustrate the case of $N=2$, while the extension to $N>2$ can be naturally handled via pairwise combinations.

\paragraph{Geometry consistency reward model}
To quantify geometric consistency between two generated sub-videos, we employ a 3D reconstruction-based reward model built upon the Pi3~\cite{wang2025pi3}.
Each video is preprocessed through the Pi3 model to obtain dense 3D point clouds $\mathcal{P}_1 \in \mathbb{R}^{N_1 \times 3}$ and $\mathcal{P}_2 \in \mathbb{R}^{N_2 \times 3}$, respectively.
Since the two videos are captured from different camera perspectives, their reconstructions reside in distinct coordinate systems.
To align them, we register $\mathcal{P}_1$ to $\mathcal{P}_2$ using Lepard~\cite{lepard2021}, a learned point cloud registration method that robustly handles partial overlaps and noisy reconstructions through keypoint-based feature matching.
After alignment, geometric similarity is quantified by the symmetric Chamfer distance:
\begin{equation}
\label{eq:geometry_reward}
\begin{split}
D_{\text{g}}(\mathcal{P}_1, \mathcal{P}_2) &= \frac{1}{2}\large[\frac{1}{N_1} \sum_{x \in \mathcal{P}_1} \min_{y \in \mathcal{P}_2} \|x - y\|_2 \\
&~~~~~~~~~~~~~ + \frac{1}{N_2}\sum_{y \in \mathcal{P}_2} \min_{x \in \mathcal{P}_1} \|y - x\|_2 \large] , \\
r_{\text{g}} &= \exp(-D_{\text{g}}) \in (0, 1],
\end{split}
\end{equation}
where $r_{\text{g}}$ is the geometry consistency reward.
As shown in~\Cref{fig:reward_score_visualize1}, lower $D_{\text{g}}$ yields higher reward $r_{\text{g}}$, encouraging the model to generate videos with consistent 3D geometry.

\begin{figure}[t]
\begin{minipage}{\linewidth}
\begin{algorithm}[H]
    \caption{\papertitle}
    \label{alg:ic_world}
    \begin{algorithmic}[1]
        \State \textbf{Input:} Initial policy video model $\pi_{\theta}$; reward scales $\lambda_\text{g}, \lambda_\text{m}$; training dataset $\mathcal{D}_\text{train}$; training timestep ratio $\tau$; denoising steps $T$; generation group size $M$; number of player $N$; learning rate $\eta$.
        \State \textbf{Output:} Video model with optimized parameters $\theta^*$.
        \For{each training step}
            \State Subsample training data $\mathcal{D}_\text{sub}$ from $\mathcal{D}_\text{train}$ ;
            \State Update old policy: $\pi_{\theta_{\text{old}}} \gets \pi_{\theta}$ ;
            \For{each $(c,\mathcal{I}) \in \mathcal{D}_\text{sub}$}
                \State \textcolor{gray}{\texttt{\# $M$ times I2V inference}}
                \State Pixel-wise couple $N$ input images $\mathcal{I}$ ;
                \State Generate $M$ samples $\{o_i\}_{i=1}^M \sim \pi_{\theta_{\text{old}}}(\cdot |c,\mathcal{I})$ ;
                \State Pixel-wise decouple $o_i$ to $N$ videos ;
                \State \textcolor{gray}{\texttt{\# Reward calculation}}
                \State Calculate $\{r_{\text{g},i}\}_{i=1}^M$ with~\Cref{eq:geometry_reward} ;
                \State Calculate $\{r_{\text{m},i}\}_{i=1}^M$ with~\Cref{eq:motion_reward} ;
                \State Rewards $\{r_{i}\}_{i=1}^M \leftarrow \{\lambda_\text{g}r_{g,i} + \lambda_\text{m} r_{m,i}\}_{i=1}^M$ ;
                \For{each sample $i \in \{1..M\}$}
                    \State Calculate advantage $A_i$ with~\Cref{eq:advantage} ;
                \EndFor
                \State \textcolor{gray}{\texttt{\# Gradient calculation}}
                \State Calculate objective $L(\theta)$ with~\Cref{eq:grpo_objective} ;
                \State Subsample $[\tau T]$ steps $T_{\text{sub}} \subset \{1..T\}$ ;
                \For{$t \in T_{\text{sub}}$}
                    \State Gradient ascent: $\theta \gets \theta + \eta \nabla_{\theta} L$ ;
                \EndFor
            \EndFor
        \EndFor
        \State Save the optimized parameters $\theta^*$.
    \end{algorithmic}
\end{algorithm}
\end{minipage}
\end{figure}


\begin{table*}[]
    \centering
    \begin{adjustbox}{width=\linewidth}
        \begin{threeparttable}
            \begin{tabular}{l|c|cccccc|cccccc}
            \toprule
            \multirow{2}{*}{Methods} & \multirow{1}{*}{Generation} & \multicolumn{6}{c|}{Static scene + Dynamic camera} & \multicolumn{6}{c}{Dynamic scene + Static camera} \\
            & time(s) & M-FID $\downarrow$ & CLIP $\uparrow$ & VLM $\uparrow$ & $\text{Geometry}_{0.1} \uparrow$ & $\text{Geometry}_{0.5} \uparrow$ & $\text{Geometry}_{0.7} \uparrow$ & M-FID $\downarrow$ & CLIP $\uparrow$ & VLM $\uparrow$ & $\text{Motion}_{10} \uparrow$ & $\text{Motion}_{20} \uparrow$ & $\text{Motion}_{30} \uparrow$ \\
            \midrule
            CogVideoX-I2V-5B & 358.50 & \textbf{\cellcolor{gray!15}96.1624} & \textbf{\cellcolor{gray!15}0.6208} & \underline{0.7684} & 0.6727 & 0.7150 & \underline{0.6909} & 202.9558 & \textbf{\cellcolor{gray!15}0.6412} & 0.7915 & 0.7853 & 0.7963 & 0.8018 \\
            Wan2.1*-14B & \underline{34.16} & 103.5042 & \underline{0.6204} & 0.7193 & 0.6898 & \underline{0.7171} & 0.6818 & \underline{185.2354} & \underline{0.6405} & \underline{0.7949} & \underline{0.8240} & \underline{0.8344} & \underline{0.8400} \\
            EasyAnimate-12B & 792.50 & 173.8106 & 0.6181 & 0.7559 & 0.6911 & 0.4775 & 0.1963 & 231.6344 & 0.6334 & 0.7719 & 0.7636 & 0.7715 & 0.7743 \\
            DynamiCrafter-1024 & 165.36 & 136.1273 & 0.6089 & 0.5609 & \underline{0.6940} & 0.2837 & 0.0654 & 216.6677 & 0.6192 & 0.7392 & 0.8101 & 0.8153 & 0.8167 \\
            VideoCrafter-1024 & 106.19 & 170.9313 & 0.6153 & 0.7656 & 0.6783 & 0.5930 & 0.3845 & 248.3784 & 0.6279 & 0.7916 & 0.7946 & 0.8036 & 0.8063 \\
            LTX-Video-13B & 89.37 & 115.3138 & 0.6170 & 0.7121 & 0.6825 & 0.7025 & 0.6393 & 200.0291 & 0.6361 & 0.7792 & 0.7957 & 0.8081 & 0.8121 \\
            \midrule
            \methodname & \textbf{\cellcolor{gray!15}17.08} & \underline{98.8324} & 0.6170 & \textbf{\cellcolor{gray!15}0.7822} & \textbf{\cellcolor{gray!15}0.6980} & \textbf{\cellcolor{gray!15}0.7217} & \textbf{\cellcolor{gray!15}0.6994} & \textbf{\cellcolor{gray!15}157.6099} & 0.6400 & \textbf{\cellcolor{gray!15}0.7958} & \textbf{\cellcolor{gray!15}0.8360} & \textbf{\cellcolor{gray!15}0.8466} & \textbf{\cellcolor{gray!15}0.8500} \\
            \bottomrule
            \end{tabular}
        \end{threeparttable}
    \end{adjustbox}
    \caption{\textbf{Shared world modeling comparison.} Generation time per video is tested on a single H20 GPU. Wan2.1* denotes the distilled version~\cite{lightx2v}. Numbers in bold indicate the best performance, while underlined numbers represent the second-best. The absolute geometry score difference appears relatively small since the inconsistency is mostly in a local region, while the metric averages over the entire generated scene and the visual difference remains significant. Please refer to Appendix for illustrative examples.}
    \label{tab:world_alignment_comparison}
\end{table*}

\begin{table*}[]
    \centering
    \begin{adjustbox}{width=0.96\linewidth}
        \begin{threeparttable}
            \begin{tabular}{lcccccccccccc}
            \toprule
            Models (I2V) & Subject Consist. & Background Consist. & Motion Smooth. & Dynamic Degree & Aesthetic Quality & Imaging Quality & Weighted Average \\
            \midrule
            \multicolumn{8}{l}{\textit{Commercial Closed Source Models.}} \\
            Gen-4 & 93.23 & 96.79 & \textbf{\cellcolor{gray!15}98.99} & \underline{55.20} & 61.77 & 70.41 & \underline{80.89} \\
            \midrule
            \multicolumn{8}{l}{\textit{Open Source Models.}} \\
            Wan2.1-14B & 94.86 & 97.07 & 97.90 & 51.38 & \underline{64.75} & 70.44 & 80.82 \\
            DynamiCrafter-1024 & \underline{95.69} & \underline{97.38} & 97.38 & 47.40 & \textbf{\cellcolor{gray!15}66.46} & 69.34 & 80.50 \\
            VideoCrafter-1024 & \textbf{\cellcolor{gray!15}97.86} & \textbf{\cellcolor{gray!15}98.79} & 98.00 & 22.60 & 60.78 & \textbf{\cellcolor{gray!15}71.68} & 78.84 \\
            CogVideoX-5B & 94.34 & 96.42 & \underline{98.40} & 33.17 & 61.87 & 70.01 & 78.61 \\
            \midrule
            \methodname & 94.22 & 95.54 & 97.26 & \textbf{\cellcolor{gray!15}72.36} & 61.05 & \underline{70.49} & \textbf{\cellcolor{gray!15}81.15} \\
            \bottomrule
            \end{tabular}
        \end{threeparttable}
    \end{adjustbox}
    \caption{\textbf{VBench metrics results.} The weighted average follows the official weighting scheme defined in VBench.}
    \label{tab:vbench}
\end{table*}

\paragraph{Motion consistency reward model}
To evaluate motion consistency between two videos, we employ a 3D point tracking-based reward model built upon SpatialTracker~\cite{xiao2025spatialtrackerv2}. 
Given two videos, we uniformly sample $T_p$ frames for each to predict camera extrinsics $\mathcal{C}_1, \mathcal{C}_2$ and 3D point trajectories $\mathcal{B}_1 = \{b_{1,i}(t_p) \in \mathbb{R}^3\}$ and $\mathcal{B}_2= \{b_{2,i}(t_p) \in \mathbb{R}^3\}, i \in [1,B], t_p \in [1,T_p]$, where each trajectory is represented in its own camera coordinate system.
We use the same definition of camera system as CameraCtrl~\cite{he2024cameractrl} and align coordinate systems of video pair according to camera extrinsics: $\mathcal{C}_a = \mathcal{C}_{1} \cdot \mathcal{C}_{2}^{-1}
$,
then align the tracks:
\begin{equation}
    \hat{b}_{1,i}(t_p) = \mathcal{C}_{a,rotate} \cdot b_{1,i}(t_p) + \mathcal{C}_{a,trans} ,
\end{equation}
where $\mathcal{C}_{a,rotate}$ and $\mathcal{C}_{a,trans} \in \mathbb{R}^3$ are the rotation and translation components of $\mathcal{C}_a$. 
We establish point correspondences by computing temporal-average positions $\bar{b}_{i} = \frac{1}{T_p}\sum_{t_p=1}^{T_p} b_{i}(t_p)$ and defining a matching function $\delta: \mathcal{B}_1 \to \mathcal{B}_2$ such that:
\begin{equation}
\delta(i) = \arg\min_{j \in [1,B]} \|\bar{b}_{1,i} - \bar{b}_{2,j}\|_2 .
\end{equation}
The motion distance is computed as the Euclidean distance:
\begin{equation}
\label{eq:motion_reward}
\begin{split}
D_{\text{m}}(\mathcal{B}_1, \mathcal{B}_2) &= \frac{1}{BT_p} \sum_{i=1}^{B} \sum_{t_p=1}^{T_p} \|\hat{b}_{1,i}(t_p) - b_{\delta(i)}(t_p)\|_2 , \\
r_{\text{m}} &= \exp(-D_{\text{m}}) ,
\end{split}
\end{equation}
where $r_{\text{m}}$ is the motion consistency reward.
As shown in~\Cref{fig:reward_score_visualize2}, higher rewards encourage the model to generate consistent 3D motion patterns across videos.

\paragraph{Algorithm framework}
Our method is outlined in~\Cref{fig:methodology} and~\Cref{alg:ic_world}.
For each text prompt and input image set $(c, \mathcal{I})$, we first combine the $N$ input images into a single large input image, then generate $M$ candidate large grid video samples. 
For each grid video sample, we split it into $N$ sub-videos, then compute the rewards with the two proposed reward models.
The final reward is the weighted average of the results of two reward models.
We use it to compute the objective and update the policy model through gradient ascent. 
This process progressively refines the model parameters $\theta$ to obtain an optimized $\theta^*$ that produces geometrically and temporally consistent videos.


\section{Experiments}


\subsection{Experimental Setup}
\label{subsec:experimental_setup}

\paragraph{Implementation details}
We employ Wan2.1-14B~\cite{wan2025} as the foundational model and we use the step-distill version~\cite{lightx2v} to initialize it, which enables us to perform 4-steps inference to generate high-quality video.
In our experiments, we consider the simplest but non-trivial setting $N=2$, where both the size of input images and output videos are 2. 
We found that LoRA finetuning~\cite{hu2022lora} produces more coherent results and a more stable training process compared with finetuning all parameters; thus, we opt to implement our algorithm based on LoRA with rank of 64.
All experiments are conducted on an 8 H20 GPUs (each with 95GB GPU memory) platform, and the finetuning of our method lasts for 40 hours with group size $M=16$ and training step up to 200.
Reward scales $\lambda_\text{g}$ and $\lambda_\text{m}$ are both set to 0.5.
We use the AdamW~\cite{loshchilov2017decoupled} optimizer with a learning rate of $1 \times 10^{-5}$.
Detailed setting of hyper-parameters can be found in Appendix.

\begin{table*}[t]
\centering

\begin{subtable}[t]{\textwidth}
\centering
\begin{adjustbox}{width=\linewidth}
\begin{tabular}{l|c|cccccc|ccccccc}
\toprule
\multirow{2}{*}{Settings} & \multirow{1}{*}{Foundation} & \multicolumn{6}{c|}{Static scene + Dynamic camera} & \multicolumn{6}{c}{Dynamic scene + Static camera} \\
& model & M-FID $\downarrow$ & CLIP $\uparrow$ & VLM $\uparrow$ & $\text{Geometry}_{0.1} \uparrow$ & $\text{Geometry}_{0.5} \uparrow$ & $\text{Geometry}_{0.7} \uparrow$ & M-FID $\downarrow$ & CLIP $\uparrow$ & VLM $\uparrow$ & $\text{Motion}_{10} \uparrow$ & $\text{Motion}_{20} \uparrow$ & $\text{Motion}_{30} \uparrow$ \\
\midrule
Zero-shot wo/ IC-generation & \multirow{2}{*}{LTX-Video-13B} & 115.3138 & 0.6170 & 0.7121 & 0.6825 & \textbf{\cellcolor{gray!15}0.7025} & 0.6393 & \textbf{\cellcolor{gray!15}200.0291} & \textbf{\cellcolor{gray!15}0.6361} & 0.7792 & 0.7957 & 0.8081 & 0.8121 \\
Zero-shot w/ IC-generation & & \textbf{\cellcolor{gray!15}94.1303} & \textbf{\cellcolor{gray!15}0.6186} & \textbf{\cellcolor{gray!15}0.7521} & \textbf{\cellcolor{gray!15}0.6845} & 0.6784 & \textbf{\cellcolor{gray!15}0.6579} & 235.7100 & 0.6328 & \textbf{\cellcolor{gray!15}0.7822} & \textbf{\cellcolor{gray!15}0.8224} & \textbf{\cellcolor{gray!15}0.8265} & \textbf{\cellcolor{gray!15}0.8280} \\
\midrule
Zero-shot wo/ IC-generation & \multirow{2}{*}{Wan2.1-14B-distill} & 103.5042 & \textbf{\cellcolor{gray!15}0.6204} & 0.7193 & 0.6898 & 0.7171 & 0.6818 & 185.2354 & \textbf{\cellcolor{gray!15}0.6405} & 0.7949 & 0.8240 & 0.8344 & 0.8400 \\
Zero-shot w/ IC-generation & & \textbf{\cellcolor{gray!15}99.1513} & 0.6174 & \textbf{\cellcolor{gray!15}0.7818} & \textbf{\cellcolor{gray!15}0.6918} & \textbf{\cellcolor{gray!15}0.7199} & \textbf{\cellcolor{gray!15}0.7006} & \textbf{\cellcolor{gray!15}160.0052} & 0.6402 & \textbf{\cellcolor{gray!15}0.7981} & \textbf{\cellcolor{gray!15}0.8342} & \textbf{\cellcolor{gray!15}0.8441} & \textbf{\cellcolor{gray!15}0.8477} \\
\bottomrule
\end{tabular}
\end{adjustbox}
\caption{\textbf{Ablation study on in-context generation} with different foundation models. The results demonstrate the effectiveness of our designs in activating the inherent in-context generation capability of large video models.}
\label{tab:ablation_study_1}
\end{subtable}

\begin{subtable}[t]{\textwidth}
\centering
\begin{adjustbox}{width=\linewidth}
\begin{tabular}{l|c|cccccc|cccccc}
\toprule
\multirow{2}{*}{Settings} & \multirow{1}{*}{Train.} & \multicolumn{6}{c|}{Static scene + Dynamic camera} & \multicolumn{6}{c}{Dynamic scene + Static camera} \\
& data & M-FID $\downarrow$ & CLIP $\uparrow$ & VLM $\uparrow$ & $\text{Geometry}_{0.1} \uparrow$ & $\text{Geometry}_{0.5} \uparrow$ & $\text{Geometry}_{0.7} \uparrow$ & M-FID $\downarrow$ & CLIP $\uparrow$ & VLM $\uparrow$ & $\text{Motion}_{10} \uparrow$ & $\text{Motion}_{20} \uparrow$ & $\text{Motion}_{30} \uparrow$ \\
\midrule
LoRA w/ SFT & 1K & 101.1831 & 0.6150 & 0.7621 & 0.6933 & 0.7079 & 0.6891 & 170.0154 & \underline{0.6400} & 0.7909 & 0.8288 & 0.8381 & 0.8426 \\
LoRA w/ SFT & 2K & 100.9812 & 0.6168 & 0.7682 & 0.6913 & 0.7121 & 0.6901 & 166.4103 & 0.6398 & \underline{0.7911} & 0.8300 & 0.8368 & 0.8447 \\
LoRA w/ SFT+GRPO & 1K & \underline{99.1012} & \textbf{\cellcolor{gray!15}0.6173} & \textbf{\cellcolor{gray!15}0.7844} & \textbf{\cellcolor{gray!15}0.7002} & \underline{0.7193} & \textbf{\cellcolor{gray!15}0.7013} & \underline{158.3412} & \textbf{\cellcolor{gray!15}0.6404} & 0.7910 & \underline{0.8310} & \textbf{\cellcolor{gray!15}0.8484} & \underline{0.8491} \\
LoRA w/ GRPO & 1K & \textbf{\cellcolor{gray!15}98.8324} & \underline{0.6170} & \underline{0.7822} & \underline{0.6980} & \textbf{\cellcolor{gray!15}0.7217} & \underline{0.6994} & \textbf{\cellcolor{gray!15}157.6099} & \underline{0.6400} & \textbf{\cellcolor{gray!15}0.7958} & \textbf{\cellcolor{gray!15}0.8360} & \underline{0.8466} & \textbf{\cellcolor{gray!15}0.8500} \\
\bottomrule
\end{tabular}
\end{adjustbox}
\caption{\textbf{Ablation study on training strategy.} All settings are based on Wan2.1-14B-distill and the IC-generation framework. For LoRA w/ SFT+GRPO, we first use SFT for warm-up training then utilize GRPO. The results show that SFT requires more data to obtain strong performance compared with GRPO. GRPO alone achieves performance comparable to SFT+GRPO while being more computationally efficient.}
\label{tab:ablation_study_2}
\end{subtable}

\begin{subtable}[t]{\textwidth}
\centering
\begin{adjustbox}{width=\linewidth}
\begin{tabular}{l|cccccc|cccccc}
\toprule
\multirow{2}{*}{Settings} & \multicolumn{6}{c|}{Static scene + Dynamic camera} & \multicolumn{6}{c}{Dynamic scene + Static camera} \\
& M-FID $\downarrow$ & CLIP $\uparrow$ & VLM $\uparrow$ & $\text{Geometry}_{0.1} \uparrow$ & $\text{Geometry}_{0.5} \uparrow$ & $\text{Geometry}_{0.7} \uparrow$ & M-FID $\downarrow$ & CLIP $\uparrow$ & VLM $\uparrow$ & $\text{Motion}_{10} \uparrow$ & $\text{Motion}_{20} \uparrow$ & $\text{Motion}_{30} \uparrow$ \\
\midrule
w/o Geometry \& Motion & 101.2243 & 0.6150 & 0.7453 & 0.6900 & 0.7180 & 0.6891 & 165.4712 & 0.6399 & 0.7926 & 0.8280 & 0.8364 & 0.8411 \\
w/ Geometry & \textbf{\cellcolor{gray!15}97.7391} & \textbf{\cellcolor{gray!15}0.6172} & \underline{0.7810} & \underline{0.6937} & \textbf{\cellcolor{gray!15}0.7270} & \textbf{\cellcolor{gray!15}0.7052} & 159.5929 & \underline{0.6401} & \underline{0.7962} & 0.8322 & 0.8380 & 0.8412 \\
w/ Motion & 99.1874 & 0.6169 & 0.7798 & 0.6898 & \underline{0.7253} & 0.6894 & \underline{159.1588} & \textbf{\cellcolor{gray!15}0.6403} & \textbf{\cellcolor{gray!15}0.7967} & \underline{0.8347} & \underline{0.8445} & \underline{0.8479} \\
w/ Geometry \& Motion & \underline{98.8324} & \underline{0.6170} & \textbf{\cellcolor{gray!15}0.7822} & \textbf{\cellcolor{gray!15}0.6980} & 0.7217 & \underline{0.6994} & \textbf{\cellcolor{gray!15}157.6099} & 0.6400 & 0.7958 & \textbf{\cellcolor{gray!15}0.8360} & \textbf{\cellcolor{gray!15}0.8466} & \textbf{\cellcolor{gray!15}0.8500} \\
\bottomrule
\end{tabular}
\end{adjustbox}
\caption{\textbf{Ablation study on reward models.} All settings are based on Wan2.1-14B-distill and the IC-generation framework. For w/o Geometry \& Motion, we use video-text alignment reward model from DanceGRPO~\cite{xue2025dancegrpo}. The results empirically demonstrate that the proposed reward models obtain substantial consistency improvement across both settings.}
\label{tab:ablation_study_3}
\end{subtable}
\caption{\textbf{Ablation study} on (a) in-context generation, (b) training strategy and (c) reward models.}
\label{tab:ablation_study}

\end{table*}




\paragraph{Datasets} 
In our experiments, we evaluate all methods under two benchmark settings.
The first setting involves static scenes with dynamic cameras, designed to assess scene-level geometric consistency.
The second setting uses dynamic scenes with static cameras, focusing on object-level motion consistency.
For the first setting, we sample 1K training and 1K testing data from DL3DV-10K~\cite{ling2024dl3dv}, a large-scale scene dataset containing 10,510 videos captured from diverse real-world locations.
For the second setting, there is currently no large-scale real-world dataset that provides multiple synchronized views of articulated objects under static cameras.
Therefore, we adopt MultiCamVideo~\cite{bai2025recammaster}, a multi-camera synchronized video dataset rendered with Unreal Engine 5.
It comprises 13.6K dynamic scenes, each recorded from 10 different camera viewpoints.
Similarly, we sample 1K training and 1K testing data from this dataset.
For all sampled data, we use Qwen2.5-VL-32B~\cite{Qwen2.5-VL} to automatically generate world descriptions, covering both static scene attributes and dynamic object behaviors.
Some examples of input data can be found in Appendix, including input images and the full corresponding prompt.

\paragraph{Baselines}
As discussed earlier, other lines of video-based world modeling works, such as multi-view video generation and 4D video generation, either rely on fixed camera trajectories or focus on object-centric generation, and their designs are highly task-specific.
Consequently, it is unfair to evaluate them under our setting, which is camera trajectory–free and not limited to object-level generation.
Therefore, we compare our approach primarily with general video-based world modeling baselines.
We establish two categories of baselines:
(1) Foundational models, including CogVideoX-5B~\cite{yang2024cogvideox}, Wan2.1-14B~\cite{wan2025}, and LTX-Video-13B~\cite{HaCohen2024LTXVideo}; and
(2) Enhanced algorithms designed to improve generation quality on foundational models, including VideoCrafter-1024~\cite{chen2023videocrafter1}, DynamiCrafter-1024~\cite{xing2023dynamicrafter}, and EasyAnimate-12B~\cite{xu2024easyanimate}.
For all baselines, we use largest available open-source versions.

\begin{figure}[t]
    \centering
    \includegraphics[width=\linewidth]{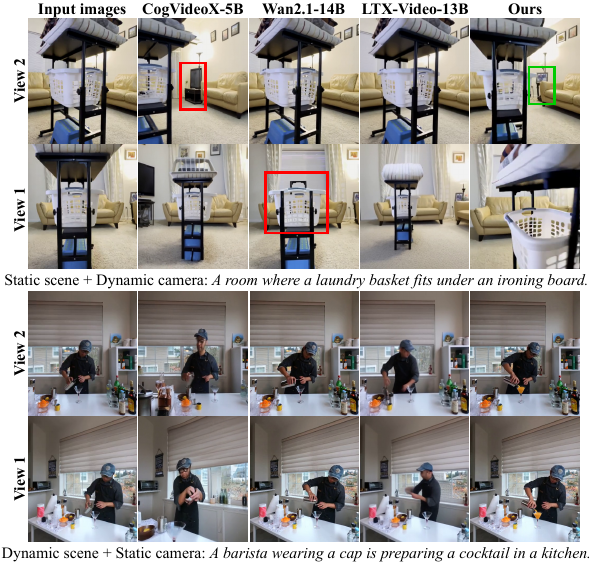}
    \caption{\textbf{Visual comparisons.} Notably, in the dynamic scene, Wan2.1 fails to maintain spatial alignment, where the man’s hands appear at a higher level in view 1 than in view 2. In contrast, our method can generate consistent results. More comparison results can be found in Appendix.}
    \label{fig:comparison}
\end{figure}

\begin{figure}[t]
    \centering
    \includegraphics[width=\linewidth]{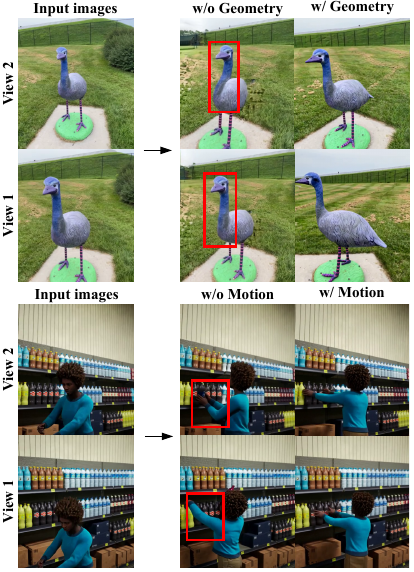}
    \caption{\textbf{Ablation on the two reward models.} We compare generations without and with the geometry reward (top) and the motion reward (bottom). Introducing the geometry reward enforces geometry consistency across views, while the motion reward improves motion consistency, reducing unnatural motion artifacts and ensuring synchronized movement.}
    \label{fig:ablation}
\end{figure}

\begin{figure*}[t]
    \centering
    \includegraphics[width=\linewidth]{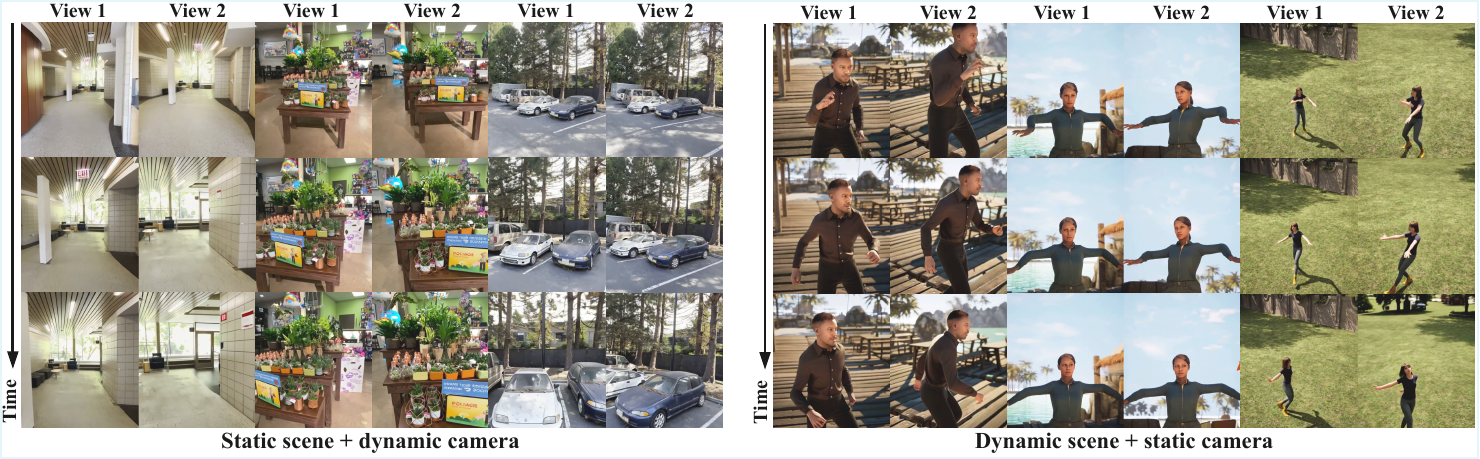}
    \caption{\textbf{Generated results visualization} on our two test sets. Our method can generate consistent and high-quality results.}
    \label{fig:results_ic_world}
\end{figure*}

\paragraph{Evaluation metrics}
\label{par:evaluation_metrics}
Evaluating the consistency of the generated shared world videos is challenging without ground truth data. 
We employ the following metrics:

\noindent\textbf{M-FID score.} For a generated set of $N$ videos, we compute the mutual Fréchet Inception Distance~\cite{heusel2017gans} (M-FID) to measure their perceptual discrepancy.

\noindent\textbf{CLIP score}~\cite{hessel2021clipscore} evaluates the similarity between the single large video, which is a pixel-wise concatenation of $N$ videos, against the text prompt.

\noindent\textbf{VLM score.} We use Qwen2.5-VL-32B~\cite{Qwen2.5-VL} to ask whether the generated $N$ videos represent the same underlying world, ranking the consistency level from 0 to 10.

\noindent\textbf{Geometry consistency score.} 
We assess geometry consistency across $N$ videos by reconstructing 3D point clouds for each video using Pi3~\cite{wang2025pi3}, then retaining only high-confidence points with three different confidence levels ($0.1, 0.5, 0.7$).
We use Lepard~\cite{lepard2021} to perform point clouds registration for a pair of videos.
After registration, we compute the Chamfer distance $D_\text{g}$ between registered point cloud pairs and the final score is reversed by $\text{exp}(-D_\text{g})$.
If $N>2$, we consider pair-wise combinations and compute the average over all the combinations as the final score.


\noindent\textbf{Motion consistency score.} 
To compute motion consistency between a pair of videos, we first uniformly sample frames at an interval of 5.
We deploy SpatialTrackerV2~\cite{xiao2025spatialtrackerv2} to predict the 3D point tracks at three density levels ($10,20,30$, with higher values indicating denser tracking). 
We align the tracks according to the two videos' camera extrinsics estimated by SpatialTrackerV2, then we calculate the Euclidean distance $D_\text{m}$ between the aligned tracks. The final score is reversed by $\text{exp}(-D_\text{m})$.
For the situation $N>2$, we calculate pairwise scores for all video combinations and report their average.


\subsection{Comparison with Existing Methods}
\label{subsec:comparison_with_existing_methods}
\vspace{-2pt}
Here we present the main comparison results, and we further present a user study in Appendix.

\paragraph{Quantitative results}
All baselines in~\Cref{tab:world_alignment_comparison} perform $N$ times sequential generation for $N$ input images.
The results show our method achieves the highest geometry and motion consistency across both settings, surpassing all baselines by clear margins.
Benefiting from our in-context generation framework,~\methodname~has the lowest generation time per video.
For broader and comprehensive comparison, we implement~\methodname~on VBench~\cite{huang2024vbench} and the results are shown in~\Cref{tab:vbench}.
All the baseline results are reported by official VBench.
\Cref{tab:vbench} shows that~\methodname~achieves an optimal weighted average score (81.15), outperforming leading closed source and open source models. 


\paragraph{Qualitative results}
We visualize our proposed~\methodname~and baseline methods in~\Cref{fig:comparison}.
In the static scene, our method accurately maintains the geometry consistency across views, with consistent object positioning and perspective (green box).
While in the dynamic scene, our method preserves the pouring action of the barista with temporal consistency from different viewpoints. More comparison results can be found in Appendix. 
Some samples of our results on the test set are shown in~\Cref{fig:results_ic_world}.



\begin{figure}[t]
    \centering
    \begin{subfigure}[t]{\linewidth}
        \centering
        \includegraphics[width=\linewidth]{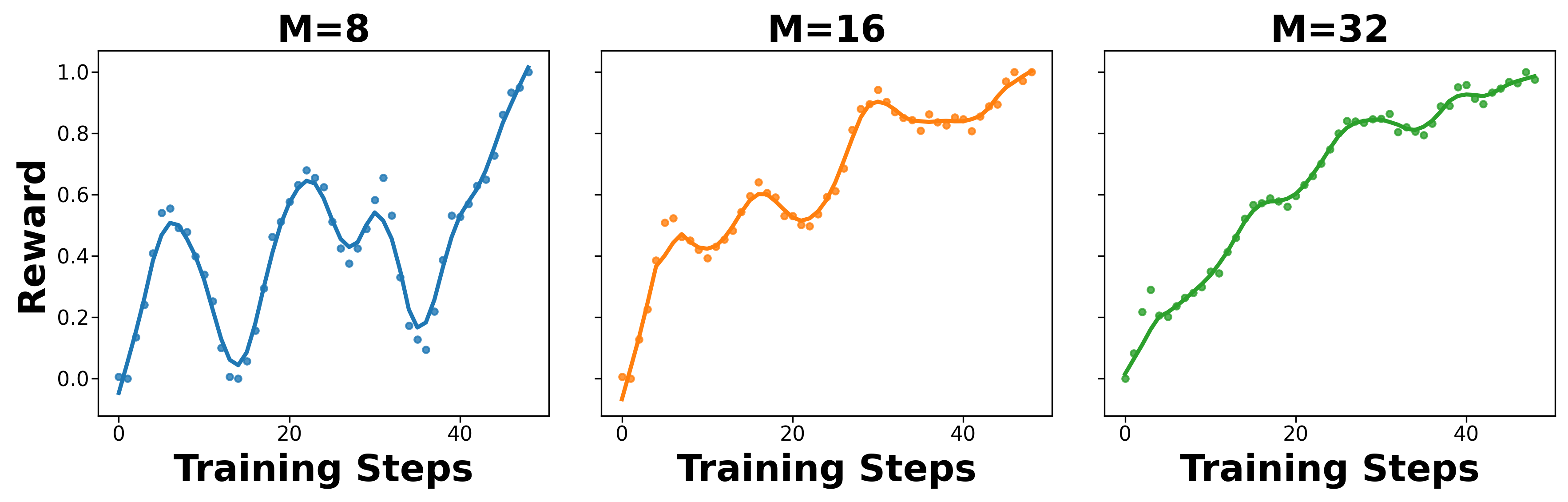}
        \caption{\textbf{Geometry consistency} reward.}
        \label{fig:reward_plot_1}
    \end{subfigure}
    
    \begin{subfigure}[t]{\linewidth}
        \centering
        \includegraphics[width=\linewidth]{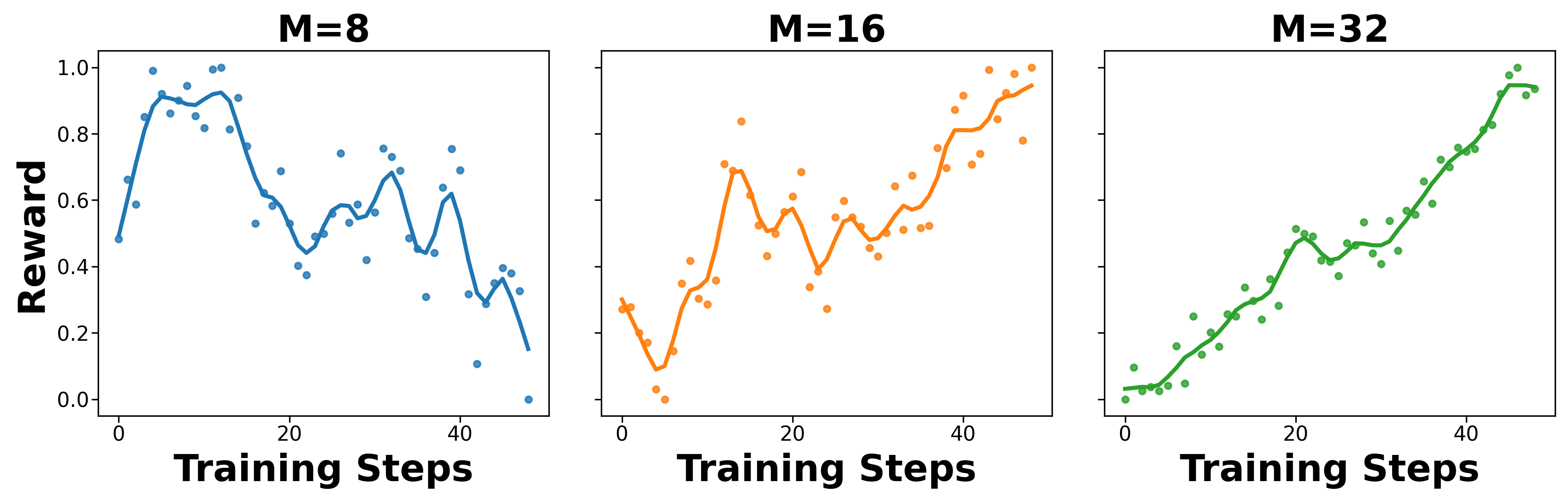}
        \caption{\textbf{Motion consistency} reward.}
        \label{fig:reward_plot_2}
    \end{subfigure}
    \caption{\textbf{Visualization of reward curves.} The generated results become more consistent with the rewards increase, and the larger group size $M$ leads to a more stable optimization process.}
    \label{fig:reward_plot}
\end{figure}

\subsection{Ablation Studies}
\label{subsec:ablation_studies}
\vspace{-2pt}

The ablation study in~\Cref{tab:ablation_study} systematically validates the contribution of each component in~\methodname.

\paragraph{In-context generation} \Cref{tab:ablation_study_1} shows that even in the zero-shot setting, our proposed in-context framework is effective in activating the inherent capability of large video models, such as LTX-Video-13B and Wan2.1-14B.

\paragraph{Training strategy} \Cref{tab:ablation_study_2} conducts an ablation study to validate our choice of training strategy for improving consistency.
The results indicate that supervised finetuning (SFT) demands substantially more data to achieve strong performance, while GRPO performs well even with limited data, making it better suited for the data-scarce shared world modeling task.
Furthermore, GRPO alone achieves performance comparable to the combined SFT+GRPO setup, justifying the use of GRPO-only fine-tuning to reduce training cost while maintaining effectiveness.


\paragraph{Reward models} \Cref{tab:ablation_study_3} and~\Cref{fig:ablation} conduct an ablation study on the two proposed reward models.
Compared with w/o geometry and motion rewards (first row), introducing the geometry reward (second row) notably improves geometry consistency (e.g., $\text{Geometry}_{0.5}$: 0.7270 vs. 0.7180) while reducing M-FID, showing that geometry consistency benefits from our explicit 3D reconstruction reward.
Similarly, adding motion reward (third row) achieves better motion consistency (e.g., $\text{Motion}_{20}$: 0.8445 vs. 0.8364).
When both geometry and motion rewards are combined (forth row),~\methodname~achieves consistent improvements across several metrics while maintaining comparable performance on the rest.
This demonstrates the general adaptability of the two proposed reward models for enhancing both geometry and motion consistency.


\subsection{Further Analysis}
\label{subsec:further_analysis}
\vspace{-3pt}

\paragraph{Visualizing the reward curves}
As shown in~\Cref{fig:reward_plot}, with the group size $M$ increasing from 8 to 32, the reward curves become smoother and more stable, indicating reduced variance and improved convergence. 
Larger groups provide more reliable relative advantage estimation, leading to steadier optimization during GRPO fine-tuning.

\begin{figure}[t]
    \centering
    \includegraphics[width=\linewidth]{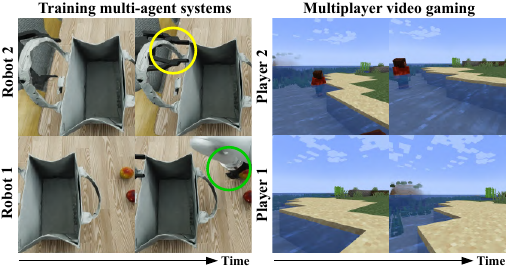}
    \caption{\textbf{Applications of~\methodname.} Our method can facilitate multiple advanced applications. \textbf{Left:} Two collaborative robots share the same workspace and coordinate to place an apple into a bag. \textbf{Right:} Two players explore the shared world in a first-person video game, where Player 1 leads while Player 2 follows.}
    \label{fig:application}
\end{figure}

\paragraph{Applications}
Unlike existing problem settings, shared world modeling requires unconstrained camera trajectories for each input image and emphasizes both object and scene-level generation.
Beyond conventional applications like multi-view video generation,~\methodname~facilitates consistent shared world modeling, empowering advanced applications such as training multi-agent systems and synchronized generation in multiplayer gaming, as shown in~\Cref{fig:application}.
 
\paragraph{Limitations}
The pixel-wise coupling operation for $N$ input images in our in-context generation framework inevitably reduces the effective resolution of each input, resulting in lower-resolution video outputs. 
A practical workaround is to apply video super-resolution techniques, such as Upscale-a-Video~\cite{zhou2024upscale}. 
Another limitation arises from the inherent capacity of current foundation models: as $N$ increases, maintaining temporal stability and cross-view coherence becomes challenging, often leading to degraded or inconsistent outputs. See Appendix for details


\vspace{-8pt}
\section{Conclusion}
\vspace{-5pt}

In this paper, we introduced~\methodname, a novel framework for shared world modeling built upon the inherent in-context generation capability of large video models. 
Unlike conventional video-based world modeling pipelines that generate views independently,~\methodname~enables parallel and coherent generation through pixel-wise coupling and instruction-based in-context prompting. 
To further enhance geometry and motion consistency, we proposed two complementary reward models built upon 3D reconstruction point clouds and 3D point trackings to optimize the framework via GRPO.
Our evaluations on newly designed benchmarks empirically demonstrate that our method substantially improves both geometry and motion consistency across the output set of videos while achieving higher efficiency than existing baselines.
Beyond surpassing prior approaches,~\methodname~opens new possibilities for advanced applications in various areas.
We hope this work could lay a solid foundation and inspire future research into shared world modeling.

  
{
    \small
    \bibliographystyle{ieeenat_fullname}
    \bibliography{main}
}

\end{document}